\documentclass[10pt,conference]{IEEEtran}
\IEEEoverridecommandlockouts
\usepackage{cite}
\usepackage{amsmath,amssymb,amsfonts}
\usepackage{graphicx}
\usepackage{textcomp}
\usepackage{xcolor}
\usepackage{hyperref}
\usepackage{booktabs}
\usepackage{multirow}
\usepackage{makecell}
\usepackage{mathtools}
\usepackage{subcaption}
\usepackage[normalem]{ulem}
\usepackage{amsfonts}
\usepackage{algorithm}
\usepackage[noend]{algpseudocode}
\usepackage{graphicx}
\usepackage{amsmath}
\usepackage{tabularx}
\usepackage{fixltx2e}
\usepackage{balance}

\usepackage[all]{xy}

\algblockdefx[Foreach]{Foreach}{EndForeach}[1]{\textbf{foreach} #1 \textbf{do}}{\textbf{end foreach}}

\newcolumntype{P}[1]{>{\centering\arraybackslash}p{#1}}

\def\BibTeX{{\rm B\kern-.05em{\sc i\kern-.025em b}\kern-.08emT\kern-.1667em\lower.7ex\hbox{E}\kern-.125emX}}

\def\syntha{\hspace*{-0.2cm}\includegraphics[scale=0.49]{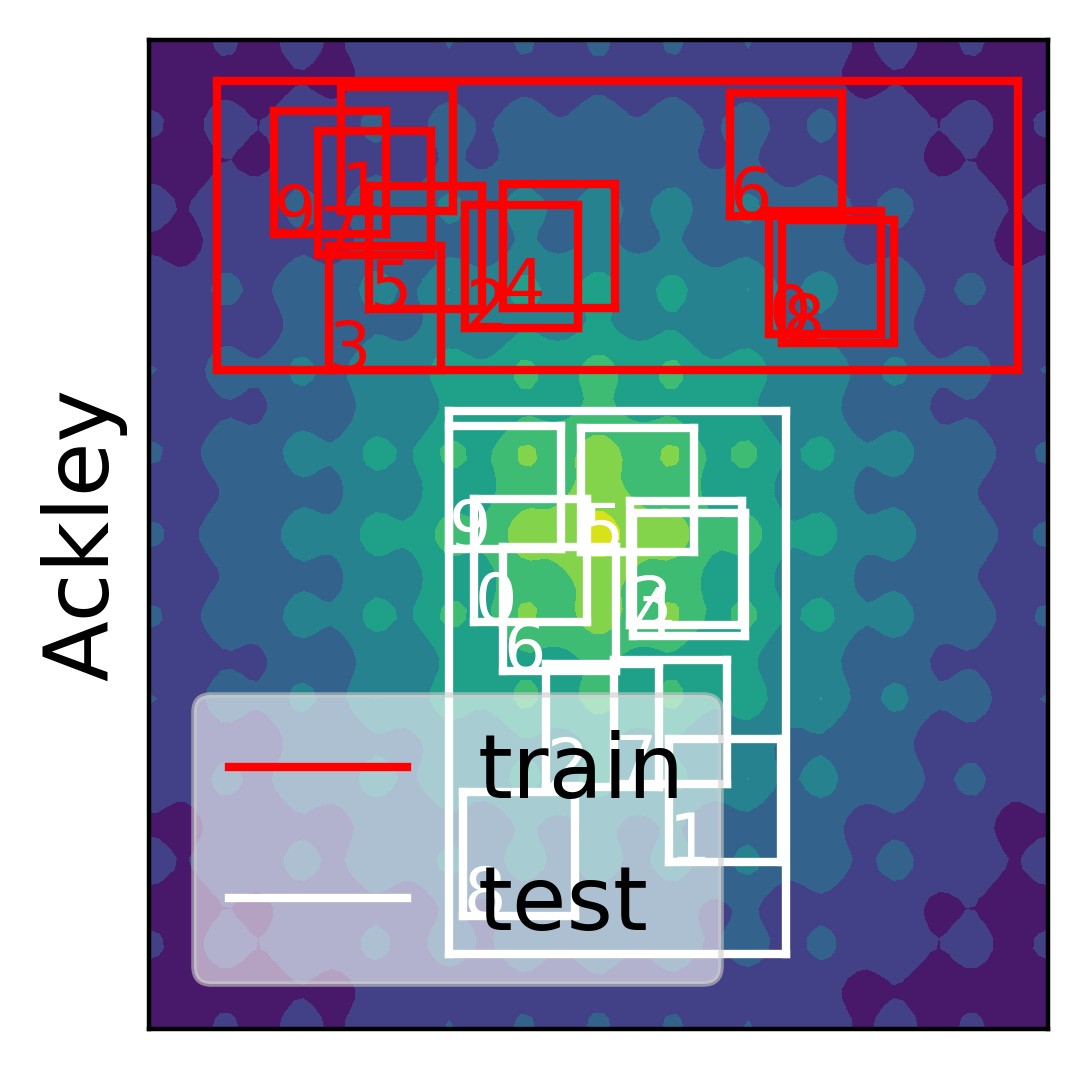}}
\def\synthb{\hspace*{-0.2cm}\includegraphics[scale=0.49]{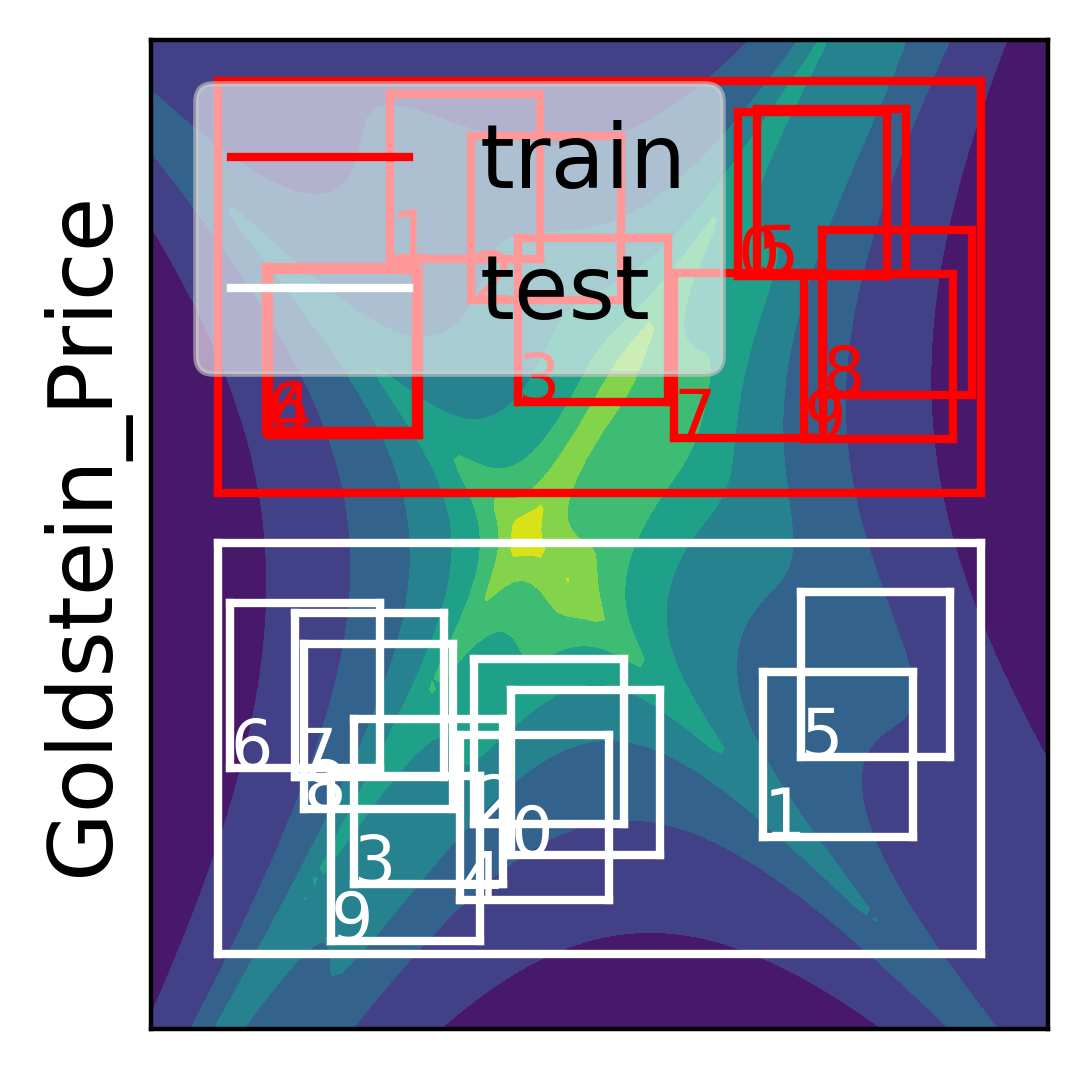}}
\def\synthc{\hspace*{-0.2cm}\includegraphics[scale=0.49]{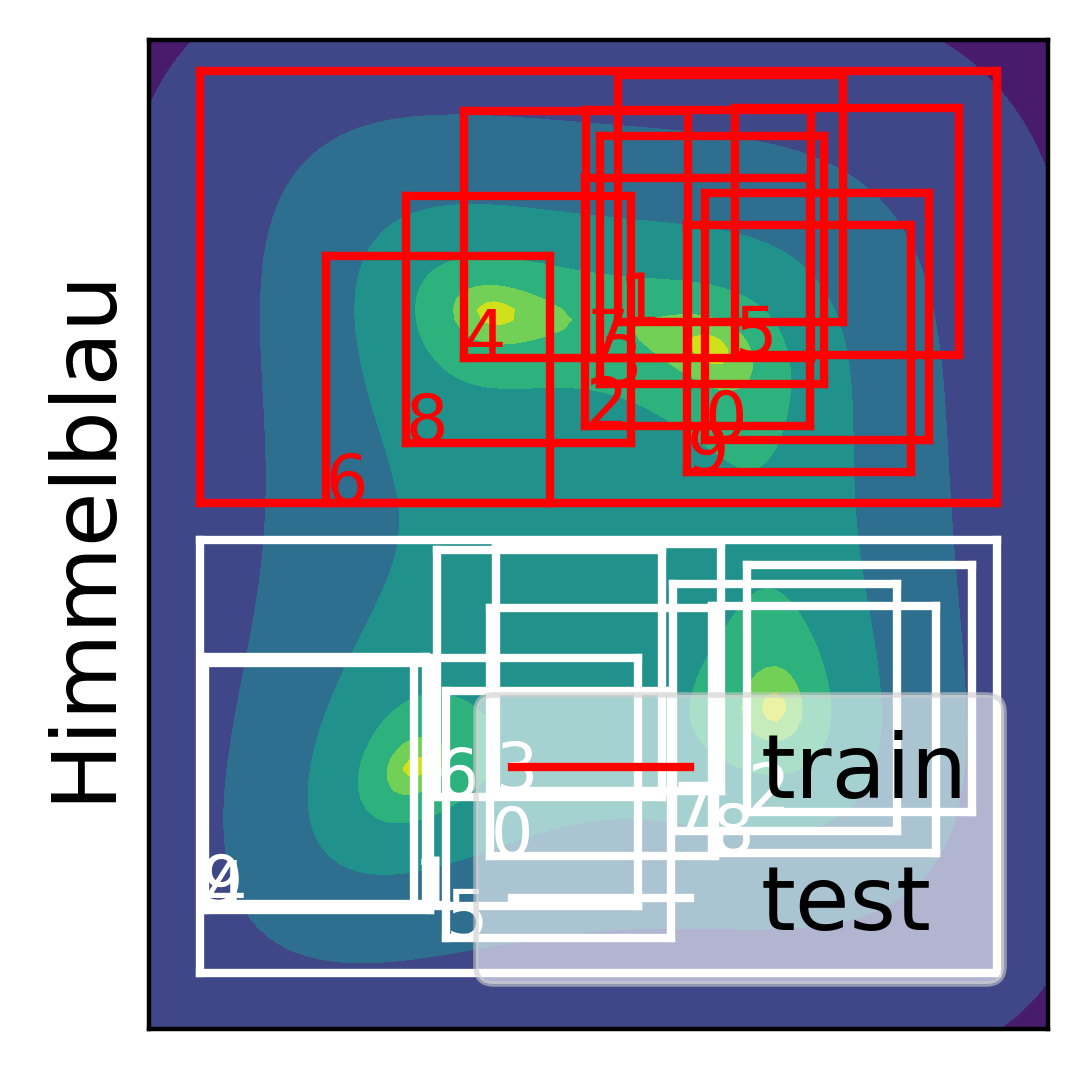}}
\def\synthd{\hspace*{-0.2cm}\includegraphics[scale=0.49]{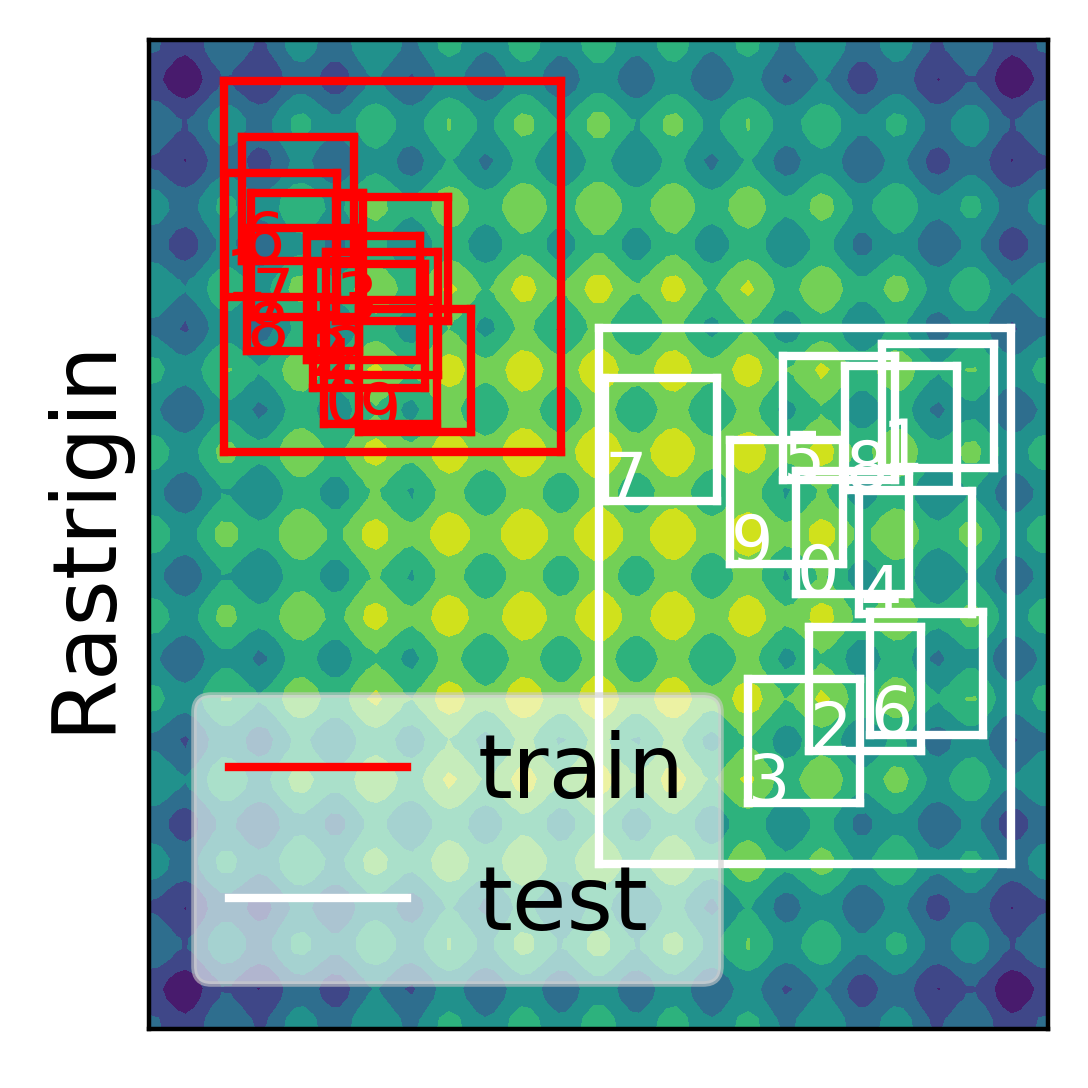}}
\def\synthe{\hspace*{-0.2cm}\includegraphics[scale=0.49]{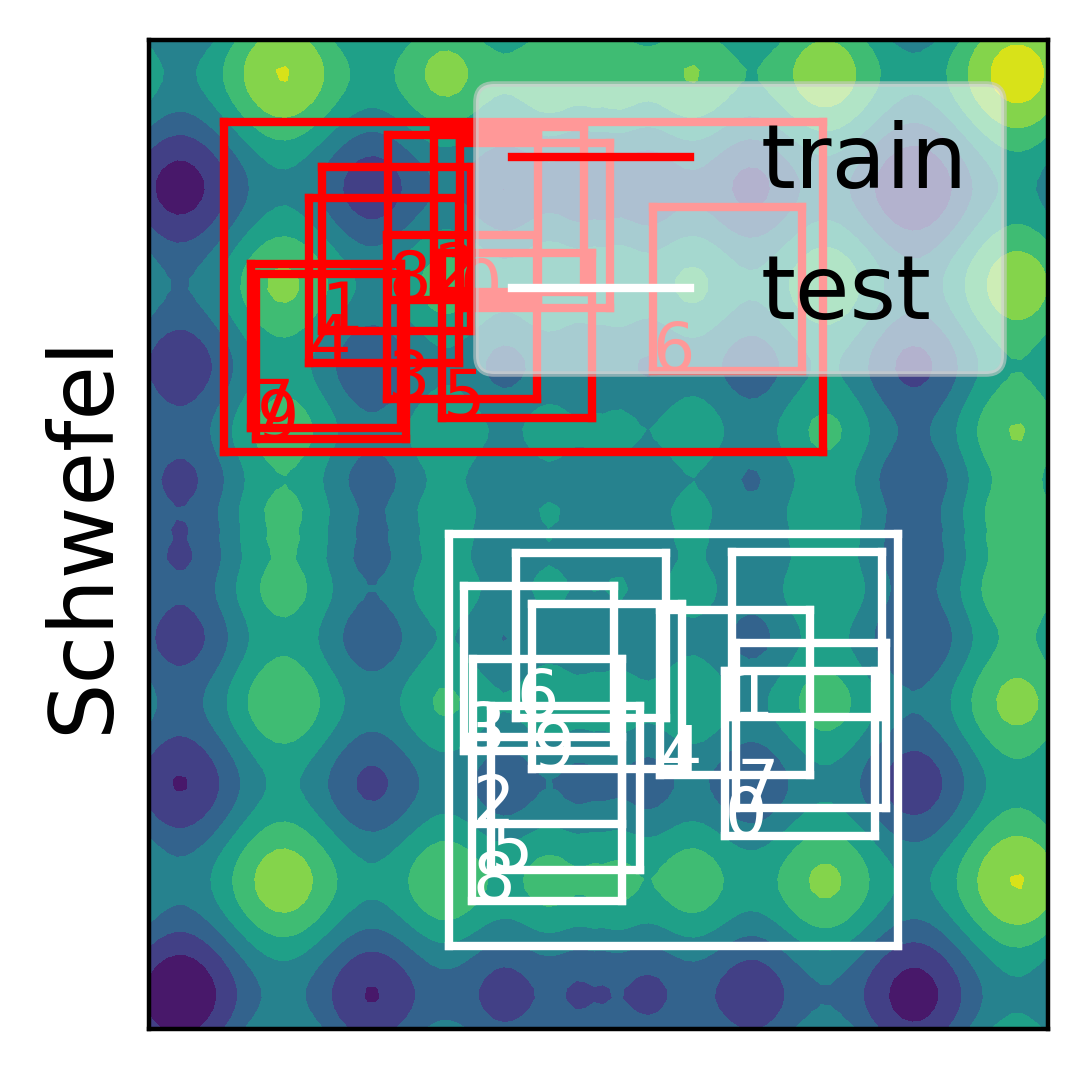}}

\usepackage{todonotes}
\newcounter{todocounter}

\makeatletter
 \newcommand{\linebreakand}{%
  \end{@IEEEauthorhalign}
  \hfill\mbox{}\par
  \mbox{}\hfill\begin{@IEEEauthorhalign}
}
\makeatother

\newcommand{\adam}[1]{\stepcounter{todocounter}
  {\color{blue!90} Adam: \thetodocounter: #1}}

\usepackage{tikz}
\newcommand*\circled[1]{\tikz[baseline=(char.base)]{
            \node[shape=circle,draw=blue,fill=cyan,text=white,inner sep=1pt] (char) {#1};}}

\begin{document}

\title{Learning to Score: Tuning Cluster Schedulers through Reinforcement Learning

}

\author{\IEEEauthorblockN{Martin Asenov} \thanks{Correspondence email: sirlab@huawei.com}
\IEEEauthorblockA{\textit{Edinburgh Research Centre} \\
\textit{Central Software Institute, Huawei}\\
0000-0003-4610-3112}
\and
\IEEEauthorblockN{Qiwen Deng}
\IEEEauthorblockA{\textit{Edinburgh Research Centre} \\
\textit{Central Software Institute, Huawei}\\
0009-0005-3663-0914}
\and
\IEEEauthorblockN{Gingfung Yeung}
\IEEEauthorblockA{\textit{Edinburgh Research Centre} \\
\textit{Central Software Institute, Huawei}\\
0000-0002-3845-0686}
\linebreakand
\IEEEauthorblockN{Adam Barker}
\IEEEauthorblockA{\textit{Edinburgh Research Centre} \\
\textit{Central Software Institute, Huawei}\\
School of Computer Science \\
University of St Andrews \\ 
0000-0002-0463-7207}
}

\maketitle

\begin{abstract}


Efficiently allocating incoming jobs to nodes in large-scale clusters can lead to substantial improvements in both cluster utilization and job performance. In order to allocate incoming jobs, cluster schedulers usually rely on a set of scoring functions to rank feasible nodes. Results from individual scoring functions are usually weighted equally, which could lead to sub-optimal deployments as the one-size-fits-all solution does not take into account the characteristics of each workload. Tuning the weights of scoring functions, however, requires expert knowledge and is computationally expensive.

This paper proposes a reinforcement learning approach for learning the weights in scheduler scoring algorithms with the overall objective of improving the end-to-end performance of jobs for a given cluster. Our approach is based on percentage improvement reward, frame-stacking, and limiting domain information. We propose a percentage improvement reward to address the objective of multi-step parameter tuning. The inclusion of frame-stacking allows for carrying information across an optimization experiment. Limiting domain information prevents overfitting and improves performance in unseen clusters and workloads. The policy is trained on different combinations of workloads and cluster setups. We demonstrate the proposed approach improves performance on average by 33\% compared to fixed weights and 12\% compared to the best-performing baseline in a lab-based serverless scenario.
\end{abstract}


\section{Introduction}

Cluster orchestration systems like Kubernetes~\cite{burns2016borg} are designed to run multiple workload types, including user-facing services, batch-processing tasks, and machine learning applications. They must carefully balance a set of competing requirements, such as ensuring high utilization at the cluster level whilst maintaining the quality of service for the underlying applications~\cite{feng2021scaling,weng2022mlaas}.

\begin{figure*}[htbp]
\includegraphics[width=1.0\linewidth]{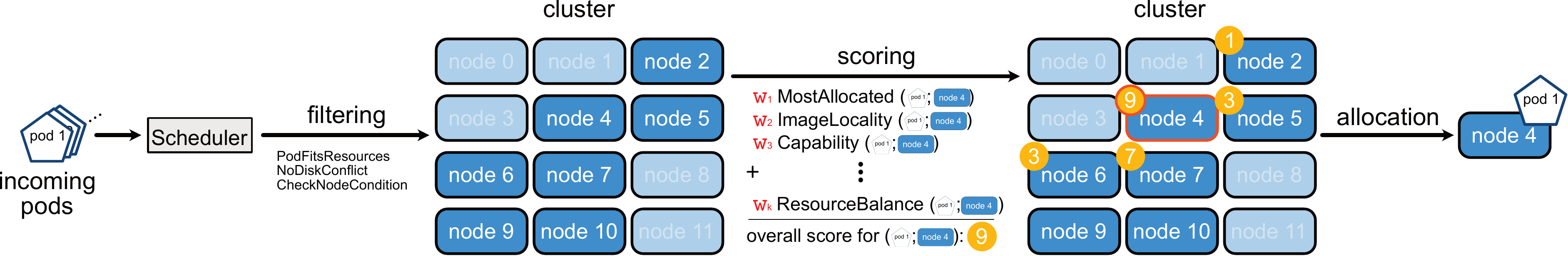}
\caption{\textbf{Filtering and scoring steps in a job scheduler.} Assigning pods to nodes in a cluster job scheduler is typically a two-step process of filtering feasible nodes, followed by scoring functions. In this work, we focus on optimizing the relative weighting (\textcolor{red}{\texttt{w}\textsubscript{1}}, \textcolor{red}{\texttt{w}\textsubscript{2}}, \textcolor{red}{\texttt{w}\textsubscript{3}}, ... ,\textcolor{red}{\texttt{w}\textsubscript{k}}) of the different scoring functions in different cluster and workload scenarios, with the goal of optimizing a given metric.}
\vspace{-1.7em}
\label{fig:scoring_overview}
\end{figure*}

One of the key tasks for the scheduler, in order to meet these requirements, is to schedule jobs (or pods in the case of Kubernetes) to nodes in the cluster. Modern cluster orchestration systems such as Kubernetes~\cite{burns2016borg}, Azure VM Allocator~\cite{hadary2020protean} and Borg~\cite{verma2015large,tirmazi2020borg} employ a two-step approach for assigning pods to nodes~\cite{wei2018research}, which is illustrated in Figure~\ref{fig:scoring_overview}. 

The first step involves selecting feasible nodes for every pod through a set of \textit{filtering} functions, which are hard constraints such as node resource capacity checks (CPU, memory, GPU) and network topology requirements, e.g., if the pod requires being in a specific region~\cite{hadary2020protean,shi2022characterizing}. The second step involves calculating scores for all feasible nodes using \textit{scoring} functions~\cite{park20183sigma,cortez2017resource,hadary2020protean}. A final score is computed by summing up the individual scores and the pod is allocated to the node with the highest normalized score.

Despite having to schedule different types of workloads with different optimization targets, scheduler scoring functions are typically weighted equally.  Specific clusters can be configured to assign different weights to prioritize certain scoring functions over others, e.g., prioritizing tighter bin packing on the cluster.  This process is, however, manual and requires knowledge of the specifics of the typical workloads, cluster configuration, and expert know-how~\cite{alabed2021high}. 

Black box optimization approaches such as random search, or Bayesian Optimization~\cite{patel2020clite,alabed2021high} can be adopted. However, tuning the weights of scoring functions is particularly difficult due to the computational cost of evaluating a new configuration. Additional challenges include the high dimensionality of workload-cluster specifications, the large number of scoring functions to be tuned and generalization to unseen configurations. 





In this paper, we propose a reinforcement learning approach to automate tuning the weights of the scoring functions to different workloads and cluster configurations. With the proposed approach, we are able to learn stronger bias for the weights sampling strategy compared to standard heuristics-based approaches. This allows us to use existing infrastructure for job scheduling while dynamically tuning the system depending on the type of workload and cluster configuration. 

Our reinforcement learning approach is based on three main ideas. First, we formulate multi-step parameter tuning as a reinforcement learning problem through the use of methods like frame stacking and techniques for balancing exploration-exploitation like entropy regularization. Second, we propose using a percentage improvement reward as the optimization target to encourage exploration. Third, we implement a simple technique to prevent overfitting and improve generalization by limiting domain information. We implement these features in a framework leveraging state-of-the-art reinforcement learning models with the option to easily add new multi-step optimization problems.

The presented approach in this paper is general, but our workloads focus primarily on serverless applications in the context of a Function as a Service (FaaS) environment, and our cluster configurations consist of heterogeneous devices ranging from powerful cloud CPU and cloud GPU machines to less powerful edge devices, which can be highly distributed.

This paper makes the following key contributions: 

\begin{itemize}
    \item Formulation of multi-step parameter tuning of weights of scoring functions as a reinforcement learning problem.
    \item Reinforcement learning approach based on percentage improvement reward, frame stacking and limiting domain information.
    \item Extensive evaluation on tuning weights of scoring functions in a FaaS system, improving performance by 33\% over constant weights, and 12\% over the best-performing optimization baseline.
\end{itemize}

The remainder of this paper is structured as follows: Section~\ref{sec:relatedwork} discusses related work. Section~\ref{sec:approach} describes our tuning approach. Section~\ref{sec:implementation} presents the system implementation and evaluation of the effectiveness of the proposed approach in a heterogeneous FaaS system. Section~\ref{sec:futurework} contains conclusions and future work.










    
    
    
\begin{figure*}[htbp]
\includegraphics[width=1.0\linewidth]{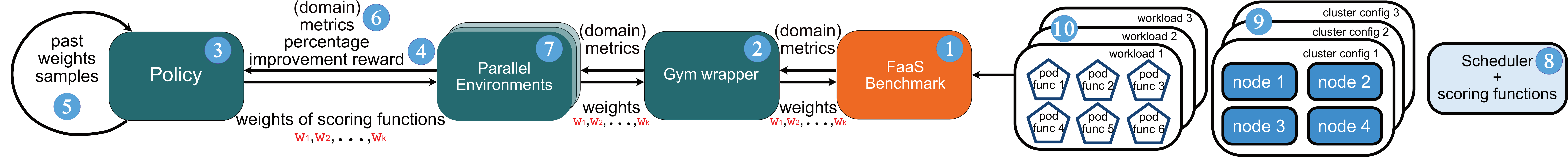}
\caption[]{\textbf{Reinforcement learning for tuning weights of scoring functions.} We pose the optimization of weights of scoring functions as a parameter tuning problem and propose a reinforcement learning based solution. In this work, we propose using percentage improvement reward~\circled{4}, encoding past samples information through the use of frame stacking or recurrent policies~\circled{5}, and limiting domain information to prevent overfitting~\circled{6}. We develop an extensive gym wrapper~\circled{2}, including the option for parallel environments~\circled{7},  and demonstrate the capability of our approach in an example FaaS benchmark scenario~\circled{1}.}
\vspace{-1.7em}
\label{fig:overview_RL}
\end{figure*}

\section{Related work}
\label{sec:relatedwork}




\subsection{Scheduling and scoring functions}
Many cluster orchestration systems employ a two-step approach of filtering and scoring for scheduling as referenced in Figure~\ref{fig:scoring_overview}. Scoring functions implement different objectives for pod-to-node allocation. For example, most-allocated and least-allocated scoring functions in Kubernetes aim for tighter packing and spreading of workloads, respectively. Alternatively, schedulers can define mathematical functions such as the piece-wise linear function that maps utilization values onto a set of defined scores to allow for more fine-grained control. In Kubernetes, this is referred to as requested-to-capacity-ratio~\cite{lee2020scheduler}. To implement specific preferences for allocation between nodes and pods, affinity and taints scoring functions are often implemented, that attract or repel a pod to a specific node~\cite{santos2019towards}. To effectively manage nodes spread around the cluster, based on region, zones, etc., topology scoring functions are implemented~\cite{verma2015large,garefalakis2018medea,hadary2020protean}. With the introduction of new scoring functions coupled with the increasing number of different workloads, weighting different scoring functions becomes an increasingly important problem. 



\subsection{Optimizing weights of scoring functions}

Different scheduling objectives are desirable, depending on the workload and cluster configurations. For example, in deep learning scenarios, we might want to pack pods in co-located nodes within the same cluster in order to achieve reduced network latency and higher throughput. Similarly, for MapReduce tasks, these tasks read data from multiple machines and have high network requirements~\cite{delay_scheduling}. On the contrary, for critical online user-facing services, we might want to spread out pods to increase redundancy due to single cluster failure.  Regardless of the high-level objective pursued, the choice of scoring functions' weights is often non-trivial. For example, within Kubernetes, efficient packing can be achieved with both MostAllocated and RequestedToCapacityRatio (RTCRatio) strategies. Moreover, scheduling them with the goal of packing can cause interference for the network and disk resources~\cite{cheng2018characterizing}. Therefore, it is useful to carefully balance the trade-off between pack and spread.

Automatically tuning the weights of the individual scoring functions is desirable to optimize for a specific target, such as application performance or bin packing of pods to nodes. This can improve the targeted metric, e.g. reducing function execution time or network traffic~\cite{rausch2021optimized}. Moreover, the optimized weights show not just a binary selection of important/insignificant scoring functions but significant relative differences, e.g. in a homogeneous cloud scenario image locality is more important than data locality~\cite{rausch2021optimized}.



\subsection{Blackbox optimization methods}

 Many blackbox optimization algorithms for parameter tuning have been proposed, from simple methods like grid and random search to more sophisticated methods that impose a different bias on the type of optimization problem that is subsequently exploited. Some of the more traditional methods include genetic algorithms (GA)~\cite{whitley1994genetic}, where based on an initial population crossover and mutation and repeated until a satisfactory result. While GA is easy to implement, it tends to scale poorly for high-dimensional problems and has no convergence guarantees. A surrogate model over the target metric could be introduced and optimized - commonly in the form of Gaussian Processes and Bayesian Optimization respectively~\cite{shahriari2015taking}. Bayesian optimizations have been applied to many aspects of systems optimization, from tuning database systems~\cite{patel2020clite}  to jobs collocation~\cite{alabed2021high}. Alternatively, optimization can be structured over the domain of parameters of interest, split below and above a certain threshold - commonly in the form of Tree-of-parzen-estimators~\cite{bergstra2011algorithms}. While those approaches impose useful biases via different kernel functions or sampling strategies via generative models of the domain variables, they tend not to take full advantage of the domain information of the targeted optimization problem.  
 Tuning weights of scoring functions is particularly challenging due to the following:
\begin{itemize}
    \item Computational cost of evaluating a new configuration
    \item High dimensionality of workload-cluster specification
    \item High dimensionality of weights of the scoring functions
\end{itemize}
As such traditional methods are ill-suited as it would take an unreasonable amount of time to converge on a desirable solution. Reinforcement learning has been presented as a viable alternative for parameter tuning in the context of parameter tuning of cluster management frameworks~\cite{karthikeyanselftune}, while also showing that it can outperform conventional approaches~\cite{jomaa2019hyp,mismar2019framework}. In this work, we expand on those approaches and address specific challenges in the context of tuning weights of scoring functions.

\subsection{Reinforcement learning for scheduling}

Reinforcement learning is defined as an optimization problem, where an agent interacts with an environment through a set of actions that change its state, with the goal of optimizing a reward~\cite{sutton2018reinforcement}.
Reinforcement learning has been applied to many diverse domains, from robot control~\cite{recht2019tour} to tuning large language models based on human preferences~\cite{openai2023gpt4}. Similarly, there has been an increasing interest from the cloud community as a viable alternative to traditional scheduling algorithms~\cite{mao2019learning,fan2021deep}. While substituting decision-making based on heuristics with an end-to-end reinforcement learning agent could lead to an impressive gain in performance, another aspect to consider is the safety aspect of a reinforcement learning agent deployed in production~\cite{amodei2016concrete}. Alternatively, reinforcement learning can be used in combination with existing infrastructure by instead tuning parameters of already deployed algorithms, such as in database systems~\cite{zhang2019end}~\cite{zhu2022magpie}. In this work, we take the latter approach and focus on tuning weights of scoring functions within job schedulers. 

\section{Approach} 
\label{sec:approach}

In this work, we focus on tuning scoring functions' weights in job schedulers in FaaS. We pose multi-step parameter tuning as a reinforcement learning problem, where we aim to achieve better sampling efficiency by learning a stronger bias from past experience. We develop a software framework for parameter tuning based on state-of-the-art reinforcement learning approaches and perform experiments with an example FaaS system. 

As seen in Figure~\ref{fig:overview_RL}, the tuning approach comprises three main components: the FaaS Benchmark~\circled{1}, Gym wrapper ~\circled{2}, and the Reinforcement learning (RL) agent~\circled{3}. The FaaS benchmark encapsulates the underlying FaaS platform for Function executions and emits metrics on how they perform. The Gym wrapper allows the FaaS benchmark to be represented as an interactive environment that takes action and emits observation, similar to OpenNetLab~\cite{eo2022opennetlab}. The agent is responsible for interacting with the FaaS environment and assigning the appropriate weights to available scoring functions. 








\subsection{Deep Reinforcement Learning Agent}

Reinforcement Learning is a machine learning approach where an agent learns how to make decisions that will lead to optimal outcomes over time.
A reinforcement learning problem is defined by a: state space - a representation of the environment at any given time; action space - a set of all possible actions the agent 
can take; and a reward function - a set of all possible rewards the agent can receive from the environment. Reinforcement learning typically uses one of two approaches: value-based or policy-based. In the value-based approach, the agent learns to estimate the value of each state-action pair and selects actions that
maximize this value. In the policy-based approach, the agent learns a policy directly without explicitly estimating the value of state-action pairs.
There is also a hybrid approach called actor-critic, which combines elements of both value-based and policy-based methods. Deep reinforcement learning refers to approaches using neural networks to represent the policy or value functions respectively. 

In this work, we leverage state-of-the-art approaches, like soft actor-critic (SAC), which evolved from max entropy reinforcement learning and the actor-critic, key idea behind is to not just maximize cumulative rewards but also make the policy more random~\cite{haarnoja2018soft} and RecurrentPPO where the PPO~\cite{schulman2017proximal} incorporating the recurrent neuron networks (RNN, LSTM, GRU) will enable the agent to handle partially observable environments better~\cite{RecurrrentPPO}. We present the training parameters in Section~\ref{sec:evaluation:training}. We formulate multi-step parameter tuning as a reinforcement learning problem. We then use this formulation to address the problem of tuning weights of scoring functions within job schedulers, with the following definition:
\begin{itemize}
  \item State:
  \begin{itemize}
    \item Static: cluster and workload information, such as number and types of machines, workload type, etc.
    \item Dynamic: encodings of action-reward pairs of explored actions so far.
  \end{itemize}
  \item Action: weights of the scoring functions.
  \item Reward: improvement over a defined metric (percentage improvement reward).
\end{itemize}


\subsubsection{Percentage Improvement Reward Function}
To encapsulate multi-step parameter tuning as a reinforcement learning objective, we propose using a percentage improvement reward~\circled{4} function defined as follows:
\vspace{-0.5em}
\begin{equation}
r_i = \begin{cases*}
  \frac{max(r_{1,},r_{2},..,r_{n})-r_{0}}{r_{0}} & if $i = n$,\\
  0,                    & otherwise.
\end{cases*}
\end{equation}
\vspace{-0.5em}

where $n$ is the number of allowed samples per experiment. This is motivated by an exploration objective, as we want the maximum percentage of improvement over a default action (e.g. same weights for all scoring functions) in one of the chosen actions across the experiment. The proposed reward has the benefits of normalization across experiments, as it is agnostic to the initial metric value from the initial action but instead optimizes the rate of improvement.

\subsubsection{Multi-step Parameter Tuning}
The state space of the reinforcement learning agent should encapsulate information about the environment. In the context of scoring functions' weights tuning, this includes
static variables such as cluster setup and workload characteristics, but also dynamic information of the performed experiments so far - pairs
of explored weights and the corresponding reward.

To encode the action-reward pairs information in addition to the static characteristics we consider two approaches~\circled{5}. The first is to present the information explicitly using frame-stacking~\cite{Mnih2015HumanlevelCT} where the number of stacks is equal to the number of maximum samples to be acquired. A second alternative is to instead use a recurrent policy such that information is encoded within the hidden state of the network.

Balancing exploration-exploitation in a systematic way is a desirable property of any parameter-tuning algorithm. Within reinforcement learning this can be achieved through simple approaches such as adding a percentage for random access, or more sophisticated approaches such as adding entropy regularization~\cite{haarnoja2018soft}.

\begin{table}[htbp]
\caption{\textbf{Skippy Scheduler Scoring Functions.} A total of eight scoring functions are used as part of scheduling. For fixed weights baseline, and initial weights selection for optimization algorithms, all weights are set to 1, except LeastAllocated and RTCRatio which are set to 0.}
\label{table:skippy_scores_func}
\footnotesize
\begin{tabularx}{\linewidth}{l|l}
    \toprule
    \textbf{Scoring Func.} & \textbf{Description} \\
    \midrule
     \texttt{LeastAllocated} & Favors nodes with the lowest utilization  \\
    \midrule
     \texttt{MostAllocated}  & Favors nodes with the highest utilization \\ 
    \midrule
     \texttt{RTCRatio} & Piecewise linear function of utilization \\ 
    \midrule
    \texttt{LocalityType} & Tag for the type of machine, e.g. edge vs cloud \\
    \midrule
    \texttt{DataLocality} & Estimated time to download necessary data \\
    \midrule
    \texttt{Capability} & Tag for capability of the machine, e.g. GPU \\
    \midrule
     \texttt{Balanced} & \multirow{2}{*}{Favors nodes with least stddev. across resources} \\ \texttt{Resource}  \\
    \midrule
    \texttt{LatencyAware} & \multirow{2}{*}{Estimated time to download the container image} \\ \texttt{ImageLocality} \\
    \bottomrule
\end{tabularx}
\vspace{-1.0em}
\end{table}

\subsubsection{Limiting Domain Information}
Generalization to unseen environments is another important property for tuning scoring functions' weights. It is desirable that once an algorithm is trained, it is able to perform well within scenarios different from the original training domain. Reinforcement learning is known for exploiting the training environment by often finding unintended shortcuts for achieving high rewards~\cite{leike2018scalable}. To mitigate the problem, we propose a simple technique of limiting static domain information to prevent overfitting and achieve better generalization~\circled{6}. This can range from including only part of the known domain information to excluding it all. This allows the algorithm to learn a good general policy for exploration and exploitation while preventing overfitting. In our experiments, we opt for the second option of limiting domain information, e.g. including only coarse description variables for workload and cluster information.

\subsection{Gym Wrapper}

The above-described contributions are implemented as a software framework for parameter tuning using reinforcement learning algorithms by developing a general environment wrapper~\circled{2}. An environment is defined by the following spaces:

\begin{itemize}
\item  Static: static parameters throughout an experiment.
\item Domain train: parameters during training.
\item Domain test: parameters for extrapolation experiments.
\item Initial action: action taken at the beginning of an experiment.
\item Reward: optimization metric of choice.
\item Actions: parameters to be optimized.
\end{itemize}

Each space described above consists of one or multiple [variable name, min, max] used for normalization purposes. The spaces, in addition to other options (e.g. hiding part of the static variables, adding noise, etc.), are then used to construct the environment. 


\section{Implementation and evaluation}
\label{sec:implementation}

To evaluate the proposed approach, we perform large-scale experiments with a high-fidelity simulator, faas-sim~\cite{rausch2021optimized}, that primarily targets the FaaS platform~\circled{1} on a set of different workloads~\circled{10} and cluster setups~\circled{9}. It captures both network topology delays, implemented with Ether~\cite{rausch2020synthesizing}, and heterogenous hardware execution time, making it suitable to evaluate scheduler placement performance in larger cluster setup~\cite{rausch2021optimized}. Skippy is the implemented scheduling system~\cite{rausch2021optimized}~\circled{8}, containing a set of scoring functions described in detail below.
\begin{figure*}[htbp]
\includegraphics[width=1.0\linewidth]{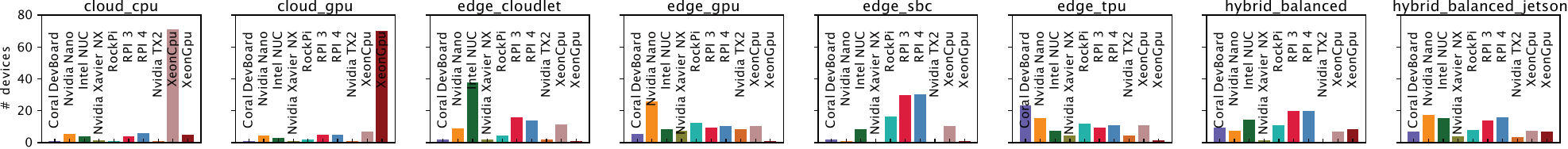}
\caption{\textbf{Different heterogeneous cluster configurations used for training and evaluation.} Distributions of the types of machines used for benchmark experiments. Only \texttt{cloud\_cpu}, \texttt{cloud\_gpu} and \texttt{edge\_cloudlet} cluster configurations are used during training. We use additional cluster configurations to evaluate how well the proposed approach is able to adapt to unseen machines` distributions.}
\vspace{-1.7em}
\label{fig:cluster_setup}
\end{figure*}
\subsection{Experimental setup}

 \paragraph{Scoring Functions} The Skippy scheduling system comes with a default scheduler. We extend the default scoring functions of Skippy with MostAllocated, LeastAllocated and RequestedToCapacityRatio inspired by their equivalents in Kubernetes as shown in Table~\ref{table:skippy_scores_func}

\begin{figure}[htbp]
\centering
\begin{subfigure}{.25\textwidth}
  \centering
  \includegraphics[width=0.9\linewidth]{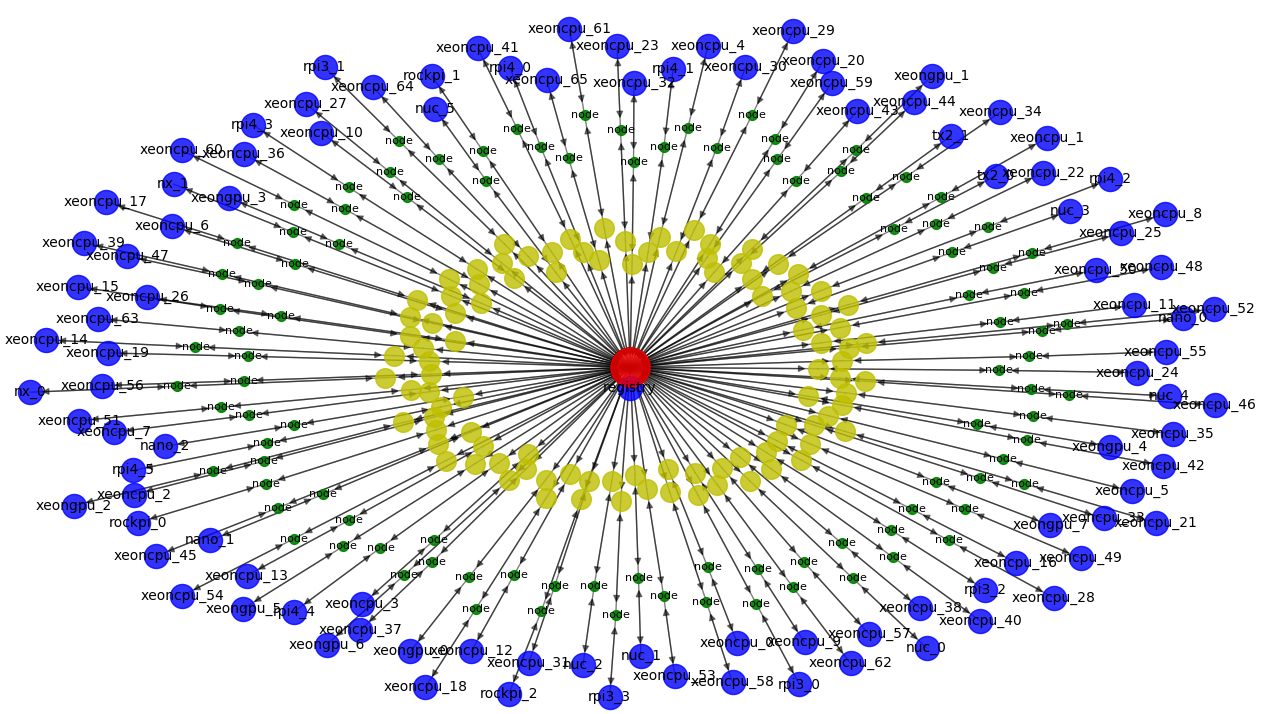}
  \caption{Internet topology}
\end{subfigure}%
\begin{subfigure}{.25\textwidth}
  \centering
  \includegraphics[width=0.9\linewidth]{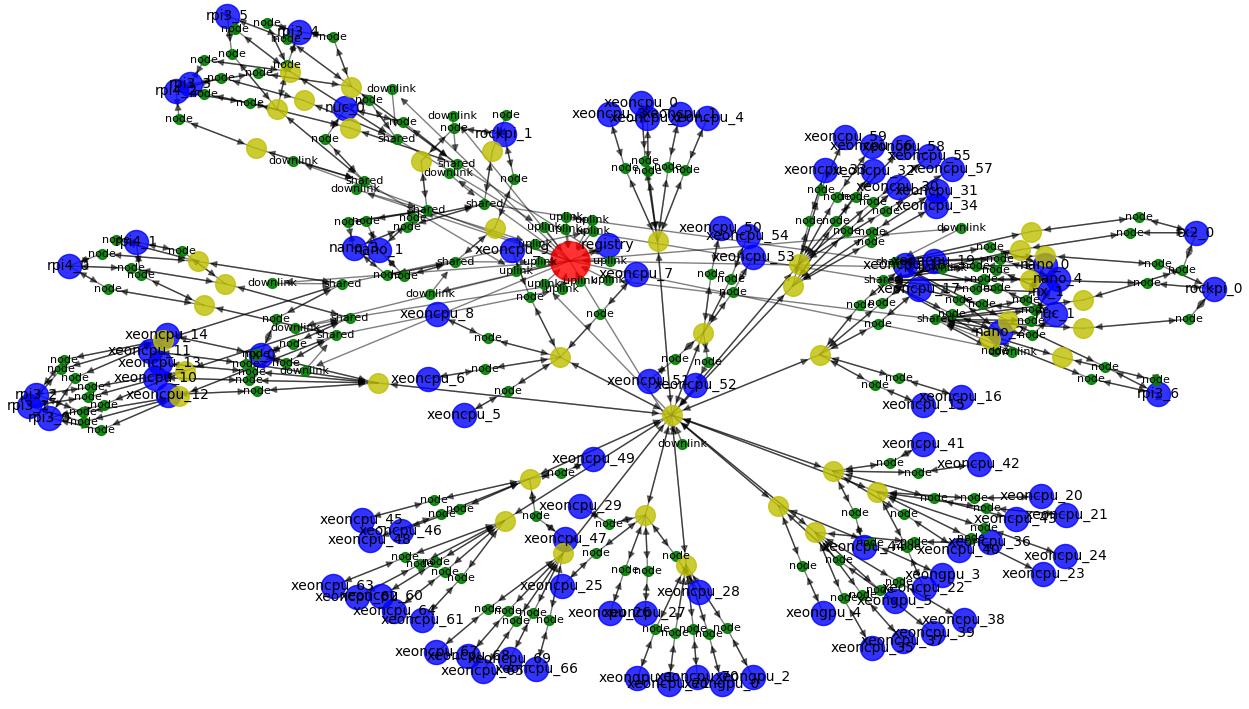}
  \caption{Urban topology}
\end{subfigure}
\caption{\textbf{Example network configurations within the cluster setup.} We use two types of cluster connectivity across benchmark experiments.}  
\vspace{-1.7em}
\label{fig:network_topology}
\end{figure}

\begin{table}[htbp]
\caption{\textbf{Cluster and workload configuration across experiments.} To evaluate how well the proposed approach generalizes to novel scenarios, we test the trained agent on novel scenarios with different cluster setups, novel workload functions, and different scheduling options. }
\label{table:train_test_setups}
\footnotesize
\begin{tabularx}{1.0\linewidth}{>{\hsize=0.53\hsize}X|>{\hsize=0.49\hsize}X|>{\hsize=0.41\hsize}X}
    \toprule
    \textbf{Configuration} & \textbf{Train environment} & \textbf{Test environment} \\
    \midrule
     Cluster setups & 3   & 8 \\
     \midrule
     Workload functions & 5   & 8  \\
     \midrule
     Requests per second & 10   &  5-30 \\
     \midrule
     \% of nodes to score & 100   &  10-100 \\
     \midrule
     Min \# nodes per func & 1  &  1-10 \\
     \midrule
     Max \# nodes per func & 100  &  50-100 \\
     \midrule
     Scale factor & 1  &  1-5 \\
     \midrule
     \# of nodes & 30-180  &  200-400 \\
    \bottomrule
\end{tabularx}
\vspace{-1.5em}

\end{table}

\paragraph{Cluster Setup} Each cluster setup consists of a variety of heterogenous hardware and network topologies. We use a total of 8 different cluster setups as defined in Figure~\ref{fig:cluster_setup}. 
\begin{itemize}
    \item \texttt{Cloud\_CPU}: configuration mainly consists of Xeon CPUs, with 71\% of total devices.
    \item \texttt{Cloud\_GPU}: configuration mainly consists of Xeon GPUs, with 70\% of total devices.
    \item \texttt{Edge\_Cloudlet}: configuration consists of a higher number of Intel NUC (mini desktop with dedicated GPU), a medium number of Raspberry PI (RPI) 3 and 4, and a lower number of NVIDIA Nano.
    \item \texttt{Edge\_GPU}: configuration mainly consists of NVIDIA Nano, and a low number of Xeon GPU. 
    \item \texttt{Edge\_SBC}: configuration mainly consists of RPI 3 and 4, and without any NVIDIA TX2 devices. 
    \item \texttt{Edge\_TPU}: configuration consists of a higher number of coral devboard and NVIDIA Nano. 
    \item \texttt{Hybrid\_balanced}: configuration consists of a similar number across devices except for NVIDIA Xavier NX and TX2. 
    \item \texttt{Hybrid\_balanced\_jetson}: have similar setup as hybrid\_balanced with higher number of NVIDIA Nano.
\end{itemize}

We use two network typologies, as shown in Figure~\ref{fig:network_topology}. 
\begin{itemize}
    \item \textit{All connected, internet topology}: where individual devices have uniform bandwidth, ensuring fast access connectivity.
    \item \textit{Limited, urban topology}: where the network is layered and has bandwidth limitation to simulate delays in connectivity.
\end{itemize}
For more extensive evaluation, we allow the disconnect between compute units and their usual network topology, i.e. we don't assume that a \texttt{cloud\_cpu} cluster configuration would necessarily have an all-connected topology. Instead, we treat cluster configurations and topology as separate factors across experiments.

\begin{figure*}[htbp]
\includegraphics[width=1.0\linewidth]{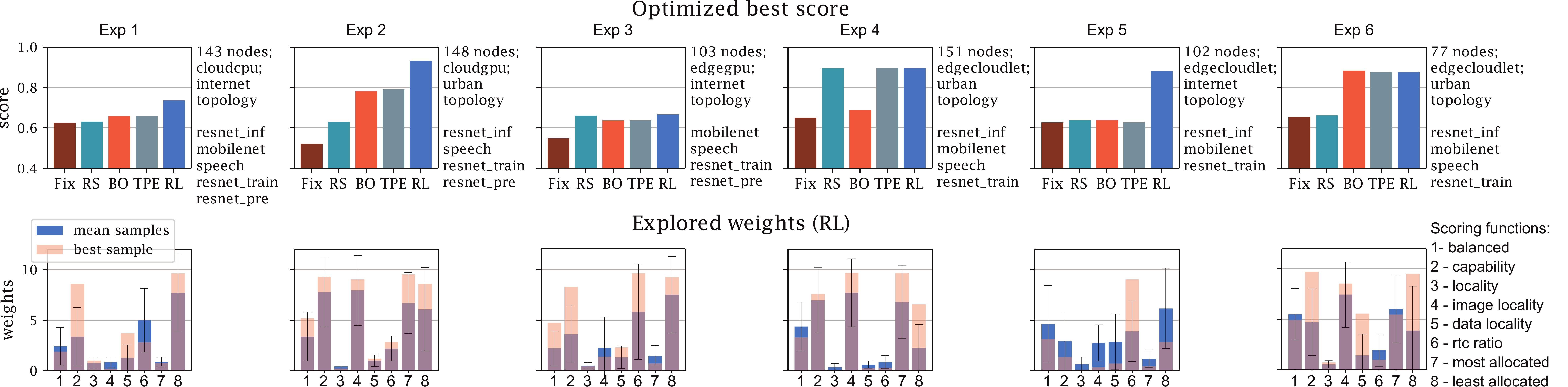}
\caption{\textbf{Tuning weights of scoring functions on \textit{similar} cluster and workload configurations.} Example results for six experiments visualized across two columns. For each experiment the following three characteristics are described (from left to right): best score (as defined in eq.~\ref{eq:score}) from the set of explored weights' configurations; mean and standard deviation, best weights selection from the reinforcement learning algorithm; short description of the experiment. We compare the proposed approach against four baselines, including fixed weights (Fix), random search (RS), Bayesian Optimization (BO), and Tree-structured Parzen Estimator (TPE). In each experiment, the fixed weight configuration was used as an initial sample (same as Fix), followed by four optimization steps. A total of eight scoring functions were used.} 
\vspace{-1.0em}
\label{fig:tuning_faas_similar}
\end{figure*}

\paragraph{Workload} We use a random combination of up to 8 different functions to form a workload. Each function follows a Poisson distribution with a constant rate of arrival. We evaluate the performance of each experiment by three metrics weighted equally - mean function execution time, mean function queue time, and the number of successful requests executed within a specified time window. Each benchmark run lasts for 100 seconds, excluding the time for the initial allocation of function pods.
\paragraph{Optimization metric}

To quantify the performance of different workloads across approaches, we use the following set of three metrics:
\begin{itemize}
\item mean function execution time: $\mu_{fet}$
\item mean function queue time: $\mu_{wait}$
\item number of successful requests executed: $N_{success}/N_{total}$
\end{itemize}

The overall \textit{score} for optimization and evaluation we define as:

\vspace{-2.0em}
\begin{multline}
 score(workload) = \\ \sum_{f\in workload}(avg(\mu_{fet}(f)+\mu_{wait}(f)+N_{success}/N_{total}))
 \label{eq:score}
\end{multline}
\vspace{-1.0em}

We normalize the three metrics between 0 and 1, so the final score is also normalized.

\begin{figure*}[htbp]
\includegraphics[width=1.0\linewidth]{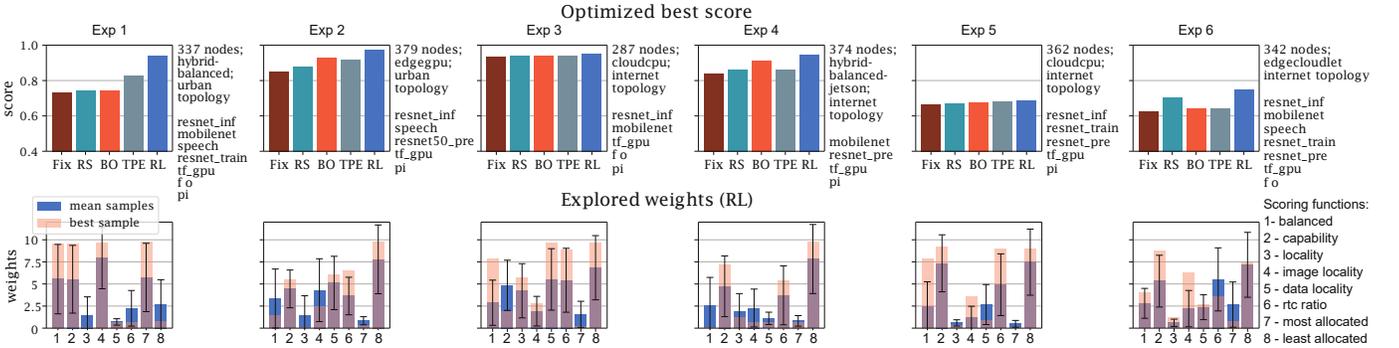}
\caption{\textbf{Example results on \textit{novel} cluster and workload configurations.} Follows the same notation as Figure~\ref{fig:tuning_faas_similar}. Results of unseen cluster and workload configurations as described in Table~\ref{table:train_test_setups}} 
\vspace{-1.5em}
\label{fig:tuning_faas_novel}
\end{figure*}

\subsection{Training}
\label{sec:evaluation:training}

For training the reinforcement learning agent, we use SAC~\cite{haarnoja2018soft} with frame stacking to account for multi-parameter tuning. Due to the explicit entropy regularization, we find that SAC achieves more robust exploration and tends not to get stuck in premature local minima during optimization. We use stable baselines\cite{raffin2021stable} with an 512x512x512 MLP network, with ReLU activations, for both QNet and policy networks. We normalize the state and action spaces and train with multiple environments in parallel. To evaluate how well the proposed method generalizes, we use just 3 out of the 8 cluster setups for training - \texttt{cloud cpu}, \texttt{cloud gpu} and \texttt{edge cloudlet}. Different workload is generated in each experiment, using a random set of 5 functions \texttt{resnet50\_training}, \texttt{resnet50\_preprocessing}, \texttt{resnet50\_inference}, \texttt{mobilenet\_inference} and \texttt{speech\_inference}. More details about training options can be seen in Table~\ref{table:train_test_setups}.

\subsection{Baselines}
\label{sec:evaluation:baselines}
To compare the proposed approach against standard methods in the literature, namely fixed scoring function weights (Fixed), random search (RS), Bayesian Optimization (BO), and Tree-Structured Parzen Estimator (TPE). Fixed weights employ a similar configuration to Kubernetes by assigning the same constant weight for all scoring functions (except LeastAllocated and RTCRatio, which have a weight of 0). Random search is one of the simplest heuristics for parameter optimization, which randomly and independently samples values for each parameter across the domain of interest. Bayesian Optimization uses a surrogate model for the underlying optimization metric, often Gaussian processes. Different biases and assumptions can then be imposed through the choice of kernel, e.g. smoothness of the underlying optimization landscapes, or in other words, similar parameter configurations lead to similar outcomes. This can then be formally posed as an optimization process through a choice of acquisition function, most of which balance between exploration and exploitation. Our experiments use a standard squared exponential kernel and an upper confidence bound acquisition function with a weight of 0.5. TPE is based on sequential model-based optimization, similar to Bayesian optimization. However, TPE utilizes a
non-parametric approach called Parzen estimators, instead of using Gaussian processes
as the surrogate model. Each optimization method uses the fixed weights as an initial sample, followed by an additional four samples. 


\subsection{Results and evaluation}

\subsubsection{Tuning scoring functions}

In the first set of experiments, we evaluate how well the proposed approach can tune the weights of the scoring functions in benchmark experiments with similar configurations, workloads, and machine distributions during training. We compare the proposed approach against baselines, as defined in Section~\ref{sec:evaluation:baselines}. For evaluation, we use ten different benchmark experiments and compare the best-obtained score (as defined in eq.\ref{eq:score}) during optimization. The best-obtained score, weights selection, and information about cluster and workload configuration in each experiment are visualized in Figure~\ref{fig:tuning_faas_similar}.




Optimizing the weights of scoring functions leads to a major increase in the above-defined metric compared to default fixed weights. The proposed approach also outperforms standard baseline approaches. We observe that in simpler experiments where the cluster has homogeneous connectivity and workload with just a few functions, tuning the weights of scoring functions does not lead to significant benefits (Exp0). However, as the number of functions in the workload grows and the cluster has heterogeneous compute units and connectivity, then optimizing the weights yields higher benefits (Exp1). We observe that the reinforcement learning agent learns a set of significant and insignificant weights - e.g. the locality has a low weights value and standard deviation across experiments, in contrast to other scoring functions such as capability. The proposed approach improves performance by 33\% over fixed weights for the scoring functions and by 20\% over the next best-performing baseline.


\begin{figure}[htbp]
\centering
\begin{subfigure}{.25\textwidth}
  \centering
  \includegraphics[width=0.90\linewidth]{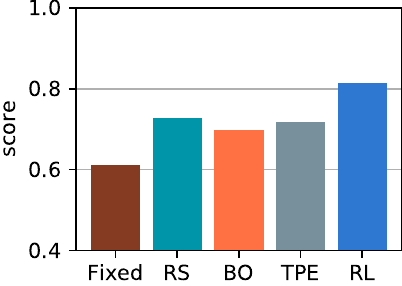}
  \caption{Similar configurations}
\end{subfigure}%
\begin{subfigure}{.25\textwidth}
  \centering
  \includegraphics[width=0.90\linewidth]{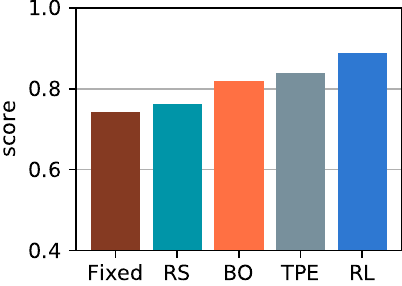}
  \caption{Novel configurations}
\end{subfigure}
\caption{\textbf{Summary results for the proposed method.} Mean of the best-achieved score (as defined in eq.~\ref{eq:score}) across ten experiments. A total of five samples of weights were used per method, with the exception of Fixed, which uses the default weights. The initial weights sample in every experiment is the same as Fixed. Configurations of the experiments in (a) and (b) are uniformly sampled from Table~\ref{table:train_test_setups}.}
\vspace{-2.0em}
\label{fig:overall_results}
\end{figure}


\subsubsection{Generalisation to other scenarios}

In this set of experiments, we evaluate how well the proposed approach can extrapolate to unseen cluster setups, workloads, and scheduling framework options. We again sample ten different configurations for evaluation but with extended workloads, additional cluster setups, and scheduling options.  Details of the differences can be seen in Table~\ref{table:train_test_setups}.


Despite very different configurations for testing, in terms of cluster, workload, and scheduling options, we observe that the proposed method again outperforms baselines as seen in Figure~\ref{fig:tuning_faas_novel}. We observe that the reinforcement learning agent is able to adapt the sampling strategy in terms of the importance of scoring functions. For example, locality weight is explored as part of the optimization in multiple experiments and has a relatively high value in Exp2 - unlike any of the experiments in Figure~\ref{fig:tuning_faas_similar}. Moreover, the overall distribution of the selected weights is often different, e.g. in Exp0 and Exp4.  In novel scenarios, the proposed approach improves performance by 20\% over fixed weights and by 6\% over the best-performing baseline. The mean score across experiments in similar and other scenarios is visualized in Figure~\ref{fig:overall_results}.

\section{Conclusions and Future work}
\label{sec:futurework}

Job scheduling in the context of ever-expanding demand for deploying heterogeneous workloads across various cluster environments remains an important consideration for maximizing efficiency. Different scoring functions often drive that process by evaluating pod-to-node allocation through desired characteristics. Yet, despite increased heterogeneity across both workloads and compute units, the scheduling process is often not tailored toward those specific needs.

In this work, we presented an approach for tuning weights of scoring functions in job schedulers using reinforcement learning. We benchmark the proposed approach on a representative FaaS scheduling system with various cluster setups and workloads. We demonstrate that the proposed approach achieves better performance in comparison to standard parameter tuning algorithms, including in scenarios that are not covered during the model training, with an improvement of up to 33\% over default static weight and up to 12\% over the best-performing baseline.

The proposed approach is well suited from an engineering standpoint as it requires minimal modification to an existing scheduling infrastructure. The proposed approach is agnostic to the number and type of scoring functions the scheduler uses. Once trained, the reinforcement learning agent can be deployed on top of an existing scheduling infrastructure with the task of tuning weights of scoring functions. Importantly, we demonstrate that the proposed method is able to generalize to unseen configurations, including different cluster setups, workloads, and scheduling options.


In future work, we will be exploring the transferability of learned policy between different scheduling systems, expanding the set of scoring functions, and using additional metrics for improved optimization.



\bibliographystyle{ieeetr}
\balance
\bibliography{ic2e2023}

\newpage

\end{document}